\documentclass[10pt, journal]{IEEEtran}
\usepackage{amsmath}
\usepackage{amssymb}
\usepackage{graphicx}

\usepackage[noend]{algpseudocode}
\usepackage{algorithm}
\DeclareMathOperator{\argmin}{arg\,min}

\title{Ranking Recovery from Limited Comparisons using Low-Rank Matrix Completion}

\usepackage{authblk}

\author[a]{Tal Levy}
\author[b]{Alireza Vahid}
\author[a]{Raja Giryes}

\affil[a]{School of Electrical Engineering, Tel-Aviv University}
\affil[b]{Electrical Engineering Department, University of Colorado Denver}

\begin{document}

% Optional adjustment to line up main text (after abstract) of first page with line numbers, when using both lineno and twocolumn options.
% You should only change this length when you've finalised the article contents.

\maketitle

\begin{abstract}
This paper proposes a new method for solving the well-known rank aggregation problem from pairwise comparisons using the method of low rank matrix completion. The 
partial and noisy data of pairwise comparisons is transformed into a matrix form. We then use tools from matrix completion, which has served as a major component
in the low-rank completion solution of the Netflix challenge, to construct the preference of the different objects. In our approach, the data of multiple comparisons is used to create
an estimate of the probability of object $i$ to win (or be chosen) over object $j$, where only a partial set of comparisons between $N$ objects is known. The data is then transformed into
a matrix form for which the noiseless solution has a known rank of one. An alternating minimization algorithm, in which the target matrix takes a {\em bilinear} form, is then
used in combination with maximum likelihood estimation for both factors. The reconstructed matrix is used to obtain the true underlying preference intensity. This work demonstrates
the improvement of our proposed algorithm over the current state-of-the-art in both simulated scenarios and real data.

\end{abstract}

\section{Introduction}
The problem of rank aggregation is common in a wide variety of tasks such as recommendation systems \cite{GroupRecommendations}, crowd sourcing \cite{CrowdSourcing}, ranking of chess players or online gamers (e.g. MSR's TrueSkill system) and many more.
In most scenarios the preference of each object (rating) is of interest as well as the global ranking of objects: Understanding the intensity of object preferences allows us to make predictions under the assumption that the preferences do not
change dramatically over time. In some scenarios, however, we are only given partial information over a collection of objects. Moreover, this information can be inconsistent due to noise. 
A common example of the problem is a small dataset of
noisy pairwise comparisons from which the preferences needs to be inferred. 

As the problem of group ranking in the presence of only partial pair-wise comparisons appears in many applications, it is of great  importance to understand the reciprocal relations between pairs that have rare or no direct interaction between them. To do so, we draw a novel link between the problem of ranking and matrix completion that allows using tools from the latter to solve problems in the former with better accuracy compared to other solutions. We demonstrate our approach on various problems including the ranking of national soccer teams showing its advantage over other approaches

This paper addresses this challenge by introducing a new algorithm based on low-rank matrix completion \cite{AltMinLRMC} in an effort to reconstruct the preference intensity. The framework of low-rank matrix completion has many powerful methods proposed for exact reconstruction from few entries \cite{LRMCwFew, SimplerLRMCApproach, SVT, PhaseTransition, HRMC}
based on convex relaxation \cite{ThePowerOfConvexRelax, MatrixComplCovexOpt} even when the entries are corrupted by noise \cite{NoisyMatComp}.
The performance of our algorithm is tested on a popular pairwise preference-based model, Bradley-Terry-Luce \cite{BTL}, and is compared with the current state-of-the-art techniques. For further
analysis, data from weather readings is used to evaluate the error on a simple pairwise partial dataset. To conclude, the method is examined on a non-trivial (complicated model) data of soccer scores from FIFA world cup, UEFA Euro and the Olympic games to create a ranking, which is then shown to be better than FIFA's men ranking in the prediction of future results as shown in Section~\ref{sec:football}.

\begin{figure*}
\begin{center}
  \includegraphics[width=0.8\textwidth]{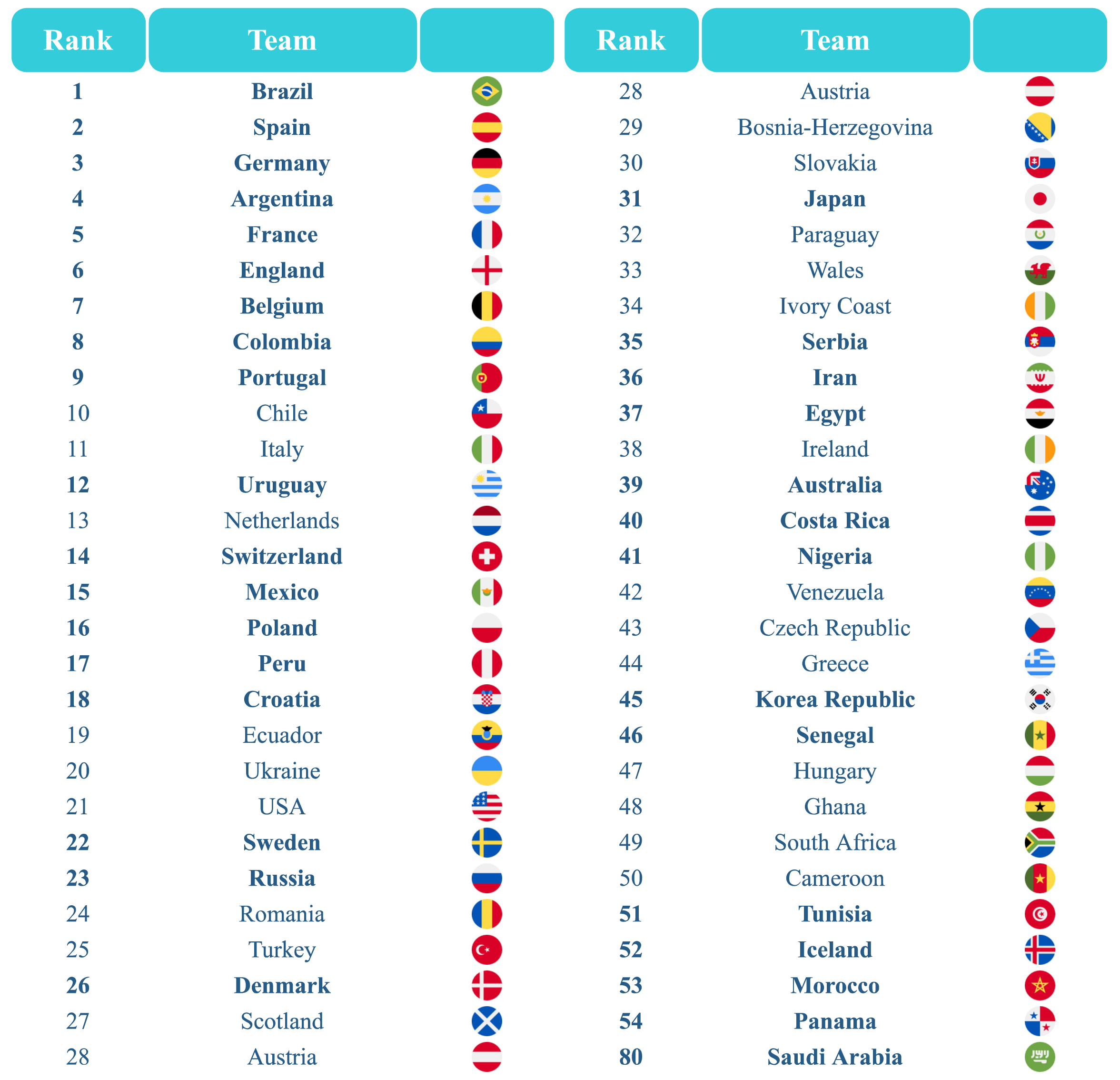}
  \end{center}
  \caption{Ranking based on all match results up to April 2018 in the past 7.5 years. In bold: Teams that qualified to 2018 FIFA World Cup.}
  \label{fig:FinalRank}
\end{figure*}

We present here a timely ranking before the 2018 FIFA World Cup tournament. 
Figure~\ref{fig:FinalRank} presents our current ranking of the top  national soccer teams including all the qualified teams to the 2018 tournament. 
This estimation is based on $7.5$ years data of all the matches between international teams up to April 2018. We used the FIFA top 100 teams as the basis for the ranking. Therefore, teams that were not ranked in the top 100 teams by FIFA on 12.04.2018 were not included. A detailed comparison to the FIFA men's ranking appears hereafter in Section~\ref{sec:football}.

\section{The Ranking Problem}
Consider the problem of rank aggregation as a simple tournament (without ties), where in each match the players (or teams) compete until one wins.
Aggregating over several past matches, part of which may repeat, provides multiple comparisons between only a subset of the possible pairs.
Assuming there is a latent preference score to the players, our task is therefore to recover a consistent ordering of all players based on the partially revealed comparison data.
Our assumption is based on the Bradley-Terry-Luce model (BTL) that postulates a set of latent scores underlying all items, where the odds of paired comparisons depend only on the relative scores of the players (or teams) involved.
%\section{The Model}

The above problem can be described by the following model.
Assume (without loss of generality) the following set of preference scores
\begin{align*}
   	\omega_1 \geq\omega_2\geq...      \geq\omega_n>0,
 \end{align*}
and a given edge set for a comparison graph:
\begin{align*}
  	 i \& j \text{ are compared  } \iff (i,j) \in \mathcal{E},
\end{align*}
where an edge is contained in the edge set $\mathcal{E}$ with some probability $p_{obs}$.
For each edge in this set, we observe $L$ repeated comparisons.
In the BTL, model the $l^{th}$ comparison between items $i$ and $j$, denoted by $y_{i/j}^{(l)}$, is
\begin{eqnarray}
   	y_{i/j}^{(l)} = 
	\begin{cases}
	1 &: \text{ w.p  } p_{i/j}=\frac{\omega_i}{\omega_i + \omega_j}  \label{eqn:Def1} \\
	0 &: \text{ otherwise  }
	\end{cases},
\end{eqnarray}
where $y_{i/j}^{(l)}$ equals 1 indicates a win for item $i$ over $j$ in the $l^{th}$ match.
In this model, it is assumed that a match result is binary, where each item either wins or loses a specific match, thus, $y_{i/j}^{(l)}=1-y_{j/i}^{(l)}$.
A naive estimator for the probability $p_{i/j}$ can be obtained by:
\begin{align}
  	y_{i/j} = \frac{1}{L}\sum_{l=1}^{L}y_{i/j}^{(l)}. \label{eqn:Def2}
\end{align}
It is clear that this is an unbiased estimator that converges to $p_{i/j}$ as $L\rightarrow\infty$. Throughout this work we will
assume that the match graph describing the comparisons (representing the edge set $\mathcal{E}$) obeys an Erd\"os R\'enyi model
$G(n,p_{obs})$, where the graph is constructed by randomly connecting nodes, with each edge having probability $p_{obs}$ of appearing
in the graph independently of the other edges.
%\section{Matrix Form}
\section{Low Rank Matrix Completion (LRMC) Approach} 
By defining the following ratio estimator, the original problem may take a matrix form:
\begin{align}
   	R_{i/j}=\frac{1}{y_{j/i}}-1. \label{eqn:Def3}
\end{align}
Notice that
\begin{align*}
	\lim_{L \rightarrow\infty}(R_{i/j})=\frac{\omega_i + \omega_j}{\omega_j}-1=\frac{\omega_i}{\omega_j}.
\end{align*}
We define the ratio matrix $M$  in the following way:
\begin{align}
   	M_{ij} =
	\begin{cases}
	R_{i/j} &: (i,j) \in \mathcal{E}\\
	0 &: \text{ otherwise  }
	\end{cases}. \label{eqn: MDef}
\end{align}
Note that all its diagonal entries are equal to one and $M_{ij}=1/{M_{ji}}$ for $(i,j) \in \mathcal{E}$.
In the case $L \rightarrow \infty$ we get the noiseless and complete matrix $\hat{M}$, which is rank-1 and can be constructed as:
\begin{align*}
\begin{split}   	
	\hat{M}&= \overrightarrow{\omega}^{T}\times\overrightarrow{\frac{1}{\omega}},\\
	\overrightarrow{\omega}&=\begin{pmatrix} \omega_1 & \omega_2 & \cdots & \omega_n \end{pmatrix},\\
	\overrightarrow{\frac{1}{\omega}}&=\begin{pmatrix} \frac{1}{\omega_1} & \frac{1}{\omega_2} & \cdots & \frac{1}{\omega_n} \end{pmatrix}.\\
\end{split}
\end{align*}
Clearly, a reconstruction of $\hat{M}$ leads also to a recovery of the ranking $\overrightarrow{\omega}$. Thus, the original problem can now be formulated as recovering $\hat{M}$ 
from a partial and noisy $M$, as this allows us to find the latent preferences $\overrightarrow{\omega}$. A suitable framework that takes advantage of the structure of the ratio matrix $\hat{M}$ to solve this problem is matrix completion.

Low rank matrix completion (LRMC) is the problem of completing a partial matrix using the lowest rank matrix fitting the observed items.
For a partially observed matrix $M$ and an observed edge set $\mathcal{E}$ the problem can be formalized as:
\begin{gather*}
   	Minimize:\text{     }Rank(X)\\
	Subject\text{ }to:\text{     }X_{ij}=M_{ij}:(i,j)\in\mathcal{E}
\end{gather*}
This is typically a non-convex problem and difficult to solve.
When the true rank is known, a simpler problem to solve is:
\begin{gather*}
   	Find\text{ }X\text{ }s.t:\text{     }Rank(X)=r\\
	Subject\text{ }to:\text{     }X_{ij}=M_{ij}:(i,j)\in\mathcal{E}
\end{gather*}
Since the target matrix $X\in \mathbb{R}^{m \times n}$ is of known rank $r$,  it can be written in a {\em bilinear} form, which will later prove to be more suited to solve. 

The matrix $X$ can be parametrized in the following form:
\begin{align*}
   	X=UV^{\dagger},
\end{align*}
where $U\in \mathbb{R}^{m \times r}$ and $V\in \mathbb{R}^{n \times r}$. 
This parameterization is common and can be found for example in sparse PCA \cite{SparsePCA} and clustering \cite{Clustering}.
In the rank recovery problem this form is advantageous  since a solution of the form $V=1 / {U}^{\dagger}$ is optimal and the rank $r$ of the matrix $M$ is equal to one.

The presented definition of the LRMC in the noiseless case for a known rank requires equality to known entries. For the ranking problem this translates to recovering $\hat{M}$ from its partial version. However, 
when $L$ is finite we have in $M$ a noisy version of the entries of $\hat{M}$. In this case a weaker condition (that is better suited for noisy data) needs to be defined in a sense of minimal error rather than equality. 
This is achieved by demanding a minimal Frobenius norm on the residual matrix (of observed entries) instead of the original equality.
By defining the operator $P_{\mathcal{E}}$
\begin{align*}
   	P_{\mathcal{E}}(X)_{ij} = 
	\begin{cases}
	X_{ij} &: (i,j)\in\mathcal{E}\\
	0 &: \text{ Else  }
	\end{cases}.
\end{align*}
$\hat{M}$ may be reconstructed by solving the following problem
\begin{align}
   	\underset{U,V}{\text{min}}(\| P_{\mathcal{E}}(UV^{\dagger})-P_{\mathcal{E}}(M) \|_{F}^2). \label{eqn: NoisyMinimizationProblem}
\end{align}
However, this results in a non-convex problem in general.

A popular approach to solve \eqref{eqn: NoisyMinimizationProblem} has been to alternately keep either $U$ or $V$ fixed and optimize over the other \cite{AltMin}.
While the overall problem is non-convex, each of these sub-problems is typically convex and can be solved efficiently.
The problem now becomes alternately solving:
\begin{gather}
   	\hat{V}^{(t)}=\underset{V}{\text{min}}(\| P_{\mathcal{E}}(\hat{U}^{(t-1)}V^{\dagger})-P_{\mathcal{E}}(M) \|_{F}^2):Given\text{ }\hat{U}^{(t-1)},  \label{eqn:Def5} \\
	\hat{U}^{(t)}=\underset{U}{\text{min}}(\| P_{\mathcal{E}}(U\hat{V}^{(t)\dagger})-P_{\mathcal{E}}(M) \|_{F}^2):Given\text{ }\hat{V}^{(t)}.   \label{eqn:Def6}
\end{gather}
A good way to initialize this process (finding $\hat{U}^{(0)}$) is to take the top-$r$ left singular vectors of $\frac{1}{p_{obs}} \cdot P_{\mathcal{E}}(M)$ using SVD \cite{AltMinLRMC}.

A solution to the minimization problem in \eqref{eqn:Def5} may be found by first defining the "row-wise" operator:
\begin{align*}
	P_{\mathcal{E}}^{(\textbf{s})}(X)_{ij} = 
	\begin{cases}
	X_{ij} &: (i,\textbf{s})\in\mathcal{E}\\
	0 &: \text{ Else  }
	\end{cases},
\end{align*}
followed by computing the following $r \times r$ matrix:
\begin{align*}
   	\hat{I}^{(s)} = P_{\mathcal{E}}^{(s)}(\hat{U})^T  P_{\mathcal{E}}^{(s)}(\hat{U}) \in \mathbb{R}^{r \times r}.
\end{align*}
Thus, the solution to \eqref{eqn:Def5} can be calculated by
\begin{align}
   	\hat{V}_{ls}^{\dagger} =((\hat{I}^{(s)})^{-1} \hat{U}^T  P_{\mathcal{E}}(M))_{ls}, \label{eqn:Sol1}
\end{align}
which has a complexity of $O( |\mathcal{E}| \cdot r^2 + n \cdot r^3)$ for $l\in[1,r]$ and $s\in[1,n]$. For the rank-1 case, equation  \eqref{eqn:Sol1} simply becomes:
\begin{align*}
   	\hat{V}_{s}^{\dagger} = \frac{\sum_{i : (i,s) \in\mathcal{E}}M_{i,s} \cdot U_{i}}{\sum_{i : (i,s) \in\mathcal{E}}U_{i}^{2}}.
\end{align*}

\section{Algorithm for Partial Noiseless Data}
For the case of $L \to \infty$ all the non-zero entries of the matrix $M$ obtained using \eqref{eqn:Def3} \& \eqref{eqn: MDef} are noiseless (identical to $\hat{M}$). Defining the observation matrix $\mathbf{Y}$ as $\mathbf{Y}_{ij}=y_{i/j}$, we present in Algorithm \ref{alg:NoiselessLRMC} a strategy to recover the ranking from $\mathbf{Y}$, the
edge set $\mathcal{E}$, the maximal weights ratio
\begin{align}
   	R_{Max}=\frac{\max(\omega)}{\min(\omega)}\equiv\frac{\omega_{Max}}{\omega_{Min}}, \label{eqn: RMaxDef}
\end{align}
and some desired resolution for the weights estimation $\Delta\omega_{Min}$.

\begin{algorithm} [H]
\caption{Noisless \textsc{LRMC} Ranking}
\label{alg:NoiselessLRMC}
\begin{algorithmic} [1]
\small
\Procedure{RankingMC}{$P_{\mathcal{E}}(\mathbf{Y}),\mathcal{E},R_{Max},\Delta\omega_{Min}$}
\State $\mathbf{Set}$: $M_{ij}=(1/\mathbf{Y}_{ji}-1)$
\State $\mathbf{Set}$: $M_{ii}=1$
\State $\mathbf{Set}$: $T = round\big(4 \cdot ln\big(n / (2\cdot\Delta\omega_{Min})\big) / ln(16)\big)$
\State $\mathbf{Set}$: $\hat{U}^{(0)}$ as the top left singular vector of $\frac{1}{p_{obs}}\cdot P_{\mathcal{E}}(M)$
\State $\mathbf{Clipping}$: set $\hat{U}^{(0)}_{i}$ where $|\hat{U}^{(0)}_{i}|>2 \cdot R_{Max}$ to zero
\State $\mathbf{Normalization}$: normalize $\hat{U}^{(0)}$ to $\hat{U}^{(0)} / \|\hat{U}^{(0)}\|_{2}$
\For{$t =1: T$}
	\State $\hat{V}^{(t)}\leftarrow \argmin_V  \big(\| P_{\mathcal{E}}(\hat{U}^{(t-1)}V^{\dagger}-M) \|_{F}^2\big)$
	\State $\hat{U}^{(t)}\leftarrow \argmin_U \big(\| P_{\mathcal{E}}(U\big(\hat{V}^{(t)}\big)^{\dagger}-M) \|_{F}^2\big)$
\EndFor
\State $\hat{U}^{(t)} = \hat{U}^{(t)} / \max_{i}\Big(\hat{U}^{(t)}_{i}\Big)$
\State \textbf{return}  $\hat{U}^{(t)}$
\EndProcedure
\normalsize
\end{algorithmic} 
\end{algorithm} 

This algorithm is based on the method presented in \cite{AltMinLRMC}.
Following the steps in the proof of theorem 2.5 in \cite{AltMinLRMC}, we get that with probability at least $1-n^{-\gamma}$ for a sampling probability ($p_{obs}$) obeying:
\begin{align}
   	p_{obs}\geq \frac{8}{3} \cdot (\gamma+1) \cdot R^{2}_{Max} \cdot \frac{ln(n)}{n} \cdot ln\bigg(\frac{n}{2\cdot\Delta\omega_{Min}}\bigg) \cdot \delta_{2}^{-2}, \label{eqn:MinpObs}
\end{align}
for some constant $\delta_{2} \leq 1/12$, we have:
\begin{align}
   	\max(|U-\hat{U}^{(t_{Final})}|) \leq \Delta\omega_{Min},
\end{align}

\section{Ranking using Limited Comparisons}

The above algorithm performs well when $L\rightarrow\infty$. However, for a finite set of comparisons $L$ it encounters a few problems.
The first is that we may get zero values in the observation matrix $\mathbf{Y}$, which will lead to infinite values in the matrix $M$. 
To solve this problem we need to truncate the values of $M$. For that purpose we use either an estimation of $R_{Max}$
or the actual value of $R_{Max}$, if it is known, to limit the value of $M$. Defining the minimal value of $y_{i/j}$ as
\begin{align}
   	y_{Min}=\frac{1}{1+C_{R}\cdot \hat{R}_{Max}},
\end{align}
the truncated observation is defined as
\begin{align}
	\hat{y}_{i/j} = 
	\begin{cases}
	y_{i/j} &:{y_{i/j}}\geq y_{Min}\\
	y_{Min} &: \text{ Else  } \label{eqn:ObservationTruncation}
	\end{cases}.
\end{align}
now we can define the truncated ratio matrix $M$ using
\begin{align}
	M_{ij} = 
	\begin{cases}
	\frac{1}{\hat{y}_{j/i}}-1 &: (i,j)\in\mathcal{E}\\
	0 &: \text{ Else  }
	\end{cases},
\end{align}
where the constant $C_{R} \geq 1$ and we use $\hat{R}_{Max}$ for the given or estimated value of $R_{Max}$.
Note that the maximal value of any entry in the matrix $M$ is now $C_{R}\cdot R_{Max}$. Though the largest value in the true matrix $\hat{M}$ cannot exceed $R_{Max}$,
in $M$ we may have several entries grater than $R_{Max}$. Because their order contains some information, we do not truncate exactly at $R_{Max}$ but rather at 
$C_{R}\cdot R_{Max}$, where $C_{R}$ is a relaxation constant. The selection of this constant trade-off the keeping of the order between those larger values and the arithmetic stability of the algorithm that is affected by the extreme values in $M$.

Another problem in Algorithm \ref{alg:NoiselessLRMC} is that for a finite $L$ the entries in $M$ become biased. Because the value of $1/\hat{y}_{i/j}$ is bounded
in the range $[1,1/y_{Min}]$ its expected value exists. Since $\phi(x)=1/x$ is a strictly convex function in the range $[y_{Min},1]$,
for a non degenerate distribution of $X$ we know from Jensen's inequality that
\begin{align*}
   	E\Big[\frac{1}{\hat{y}_{i/j}}\Big] > \frac{1}{E[\hat{y}_{i/j}]}.
\end{align*}
Thus, from this inequality we have
\begin{align*}
   	E[M_{ji}]=E\Big[\frac{1}{\hat{y}_{i/j}}\Big]-1 > \frac{1}{E[\hat{y}_{i/j}]} - 1.
\end{align*}
A second bias factor comes from the truncation and thus
\begin{align*}
   	\frac{1}{E[\hat{y}_{i/j}]} < \frac{1}{E[y_{i/j}]} = \frac{1}{p_{ij} }.
\end{align*}
Even though the two bias factors are opposite to each other they will rarely cancel each other, so the expected value
$E[M_{ij}]$ will most likely remain biased. 
Minimizing a squared error function such as the Frobenius norm on an estimated matrix $M$, which 
is a biased estimator for the true matrix $\hat{M}$ with  partially observed entries, will result
in a biased estimation of the ratios $\omega_{i} / \omega_{j}$.

The third problem in Algorithm \ref{alg:NoiselessLRMC} is the different variance of different entries in $M$, which strongly depends
on the probabilities $p_{i/j}$. To resolve the last two problems we replace the Frobenius norm minimization in \eqref{eqn:Def5} and \eqref{eqn:Def6} with the
maximum likelihood estimator for $U$ and $V$ given $M,\mathcal{E}$ and the
result obtained from the previous iteration. Since each pair may be compared a different number of times against each other, in the analysis we will replace the number of comparisons $L$ with a matrix $\mathbf{L}$ containing 
the number of comparison per each pair. The likelihood function is given by
\begin{align}
	\ell(\hat{V}^{(t)}|\mathbf{L},\mathbf{Y},\hat{U}^{(t-1)})=\prod_{(i,j)\in\mathcal{E}} {{L_{ij}}\choose{k_{ij}}} p_{i/j}^{k_{ij}}(1-p_{i/j})^{L_{ij}-k_{ij}}, \label{eqn: LikelihoodFunction}
\end{align}
where $p_{i/j}=1/(1+\hat{M}_{ji})$, $\hat{M}_{ij}=U_{i}V_{j}$ and $k_{ij}=L_{ij}\cdot\mathbf{Y}_{ij}$.
For a given factor $\hat{U}^{(t-1)}$ that approximates $U$, we may define $p_{i/j}$ in terms of $V^{(t)}$ using the following
\begin{align}
	p_{i/j}=\frac{1}{1+\hat{U}^{(t-1)}_{j}\hat{V}_{i}^{(t)}}. \label{eqn: pFromUandV}
\end{align}
Plugging \eqref{eqn: pFromUandV} to \eqref{eqn: LikelihoodFunction} leads to the following log-likelihood objective function (without terms independent of $V$)

\footnotesize
\begin{align*}
	\mathcal{L}=\sum_{(i,j)\in\mathcal{E}}(L_{ij}-k_{ij})\Big[\log\Big(\hat{U}^{(t-1)}_{j}\hat{V}_{i}^{(t)}\Big)\Big]-L_{ij}\log\Big(1+\hat{U}^{(t-1)}_{j}\hat{V}_{i}^{(t)}\Big).
\end{align*}
\normalsize

\noindent To maximize $\mathcal{L}$ with respect to $\hat{V}^{(t)}_{q}$, we take the derivative
\begin{align}
	\frac{d\mathcal{L}}{d\hat{V}^{(t)}_{q}}=\sum_{j:(q,j)\in\mathcal{E}}\Bigg(\frac{L_{qj}-k_{qj}}{\hat{V}^{(t)}_{q}}-\frac{L_{qj}\cdot\hat{U}^{(t-1)}_{j}}{1+\hat{U}^{(t-1)}_{j}\hat{V}_{q}^{(t)}}\Bigg)=0. \label{eqn:MaxLogL}
\end{align}
By some arithmetical operations, \eqref{eqn:MaxLogL} becomes
\begin{align}
	\frac{1}{\hat{V}^{(t)}_{q}}\cdot\sum_{j:(q,j)\in\mathcal{E}}\Bigg(\frac{L_{qj}}{1+\hat{U}^{(t-1)}_{j}\hat{V}_{q}^{(t)}}-k_{qj}\Bigg)=0.
\end{align}
Since $\hat{V}^{(t)}_{q} > 0$, we obtain the following equation
\begin{align}
	\sum_{j:(q,j)\in\mathcal{E}}\frac{L_{qj}}{1+\hat{U}^{(t-1)}_{j}\hat{V}_{q}^{(t)}}=\sum_{j:(q,j)\in\mathcal{E}}L_{qj}\cdot\frac{k_{qj}}{L_{qj}}.
\end{align}
Dividing both sides by $n$ and assigning $k_{qj}=L_{qj}\cdot y_{q/j}$
\begin{align}
	\frac{1}{n}\cdot\sum_{j:(q,j)\in\mathcal{E}}\frac{L_{qj}}{1+\hat{U}^{(t-1)}_{j}\hat{V}_{q}^{(t)}}=\frac{1}{n}\cdot\sum_{j:(q,j)\in\mathcal{E}}L_{qj}\cdot y_{q/j},
\end{align}
or in terms of the ratio matrix $M$
\begin{align}
	\frac{1}{n}\cdot\sum_{j:(q,j)\in\mathcal{E}}\frac{L_{qj}}{1+\hat{U}^{(t-1)}_{j}\hat{V}_{q}^{(t)}}=\frac{1}{n}\cdot\sum_{j:(q,j)\in\mathcal{E}}\frac{L_{qj}}{1+M_{jq}}. \label{eqn: LikelihoodSolution1}
\end{align}
Defining the weighting factor
\begin{align}
	W_{qj} = \frac{L_{qj}}{\max_{j}(L_{qj})}, \label{eqn: WeightSolution1}
\end{align}
and deviding \ref{eqn: LikelihoodSolution1} by $\max_{j}(L_{qj})$ we have:
\begin{align}
	\frac{1}{n}\cdot\sum_{j:(q,j)\in\mathcal{E}}\frac{W_{qj}}{1+\hat{U}^{(t-1)}_{j}\hat{V}_{q}^{(t)}}=\frac{1}{n}\cdot\sum_{j:(q,j)\in\mathcal{E}}\frac{W_{qj}}{1+M_{jq}}. \label{eqn: LikelihoodSolution}
\end{align}
By defining the constant
\begin{align}
	 S_{:q} \equiv \frac{1}{n}\cdot\sum_{j:(q,j)\in\mathcal{E}}\frac{W_{qj}}{1+M_{jq}} ,
\end{align}
and the transformation
\begin{align*}
	Z \equiv \frac{1}{\hat{V}_{q}^{(t)}},
\end{align*}
where both $Z$ and $S_{:q}$ are in the range $[0,1]$, \eqref{eqn: LikelihoodSolution} becomes:
\begin{align}
	f(Z) = \frac{1}{n}\cdot\sum_{j:(q,j)\in\mathcal{E}}\frac{W_{qj}}{1+\frac{\hat{U}^{(t-1)}_{j}}{Z}}=S_{:q}.
\end{align}
Note that the maximal valid value of $f(Z)$, (when $Z=1$) is
\begin{align}
	\max_{Z}(f(Z))=\frac{1}{n}\cdot\sum_{j:(q,j)\in\mathcal{E}}\frac{W_{qj}}{1+\hat{U}^{(t-1)}_{j}} \label{eqn: MaxFZ}.
\end{align}
Therefore, if the calculated constant $S_{:q}$ is larger than \eqref{eqn: MaxFZ} we return $Z_{0}=1$, since values of $Z$ grater than one are not valid.
Otherwise, we use the fact that the $f(Z)$ is a strictly monotone function in the range $Z\in [0,1]$ and define the function
\begin{align}
	g(Z) = f(Z) - S_{:q},
\end{align}
which is also strictly monotone and has one root in the interval $Z\in [0,1]$. The root can be easily
found using either the simple bisection method or the faster Brent's method.
Even for the simple bisection and a required accuracy $\delta_{Z}$, we will find a good approximation for the root in $O(log_{2}(1/\delta_{Z}))$ steps.
The solution for $\hat{V}_{q}^{(t)}$ in this case is obtained using the root approximation ($Z_{0}$) for the function  $g(Z)$ as $\hat{V}_{q}^{(t)}=1/Z_{0}$.

We will define the process of calculating the MLE of $\hat{V}_{q}^{(t)}$  as
\begin{align*}
	\hat{V}_{q}^{(t)}=FactorMLE(M,\mathcal{E},\hat{U}^{(t-1)},\delta_{Z},\mathbf{L}).
\end{align*}
For the MLE of $\hat{U}_{q}^{(t)}$ we define
\begin{align}
	S_{q:} \equiv \frac{1}{n}\cdot\sum_{i:(i,q)\in\mathcal{E}}\frac{W_{iq}}{1+\hat{U}^{(t)}_{q}\hat{V}_{i}^{(t)}},
\end{align}
where $W_{iq}$ is as defined in \ref{eqn: WeightSolution1}.
Now we have to solve
\begin{align}
	f(Z) = \frac{1}{n}\cdot\sum_{i:(i,q)\in\mathcal{E}}\frac{W_{iq}}{1+Z\hat{V}_{i}^{(t)}}=S_{q:}.
\end{align}
In this case, the minimal value of $f(Z)$ is
\begin{align}
	\min_{Z}(f(Z))=\frac{1}{n}\cdot\sum_{j:(q,j)\in\mathcal{E}}\frac{W_{iq}}{1+\hat{V}^{(t)}_{j}},
\end{align}
obtained for $Z=1$. Thus, if we have
\begin{align}
	S_{q:} \leq \frac{1}{n}\cdot\sum_{i:(i,q)\in\mathcal{E}}\frac{W_{iq}}{1+\hat{V}_{i}^{(t)}},
\end{align}
we will assign $\hat{U}_{q}^{(t)}=1$. Otherwise, we use the fact that $f(Z)$ is strictly monotone to get a numerical solution by the method described 
above for $\hat{V}_{q}^{(t)}$, we define the entire process as
\begin{align*}
	\hat{U}_{q}^{(t)}=FactorMLE(M,\mathcal{E},\hat{V}^{(t)},\delta_{Z},\mathbf{L}).
\end{align*}
For low values in $\mathbf{L}$, the entries of $M$ are very noisy and may get extreme values. To avoid
such values from affecting the result, we propose to truncate the values of $U$ and $V$ at each iteration. The values to be truncated are the values of $\hat{U}$, which estimates $\overrightarrow{\omega}$, that are smaller than $1/(C_{R}\cdot R_{Max})$,
and the value of  $\hat{V}$, which estimates $\overrightarrow{1/\omega}$, that are larger than $C_{R}\cdot R_{Max}$.

Since we know that for the optimal solution we have $U_{q} = 1/V_{q}$, we can force the consistency
of the solution for $U$ and $V$ at each iteration with the optimal solution by applying the steps
\begin{align}
\begin{split}
   	\hat{R}_{q}^{(t)} &= \frac{\hat{U}_{q}^{(t)}+\frac{1}{\hat{V}_{q}^{(t)}}}{2},\\
	\hat{U}_{q}^{(t)} &= \hat{R}_{q}^{(t)},\\ \label{eqn:Consistency}
	\hat{V}_{q}^{(t)} &=\frac{1}{\hat{R}_{q}^{(t)}}.
\end{split}
\end{align}
We will define the set of assignments in \eqref{eqn:Consistency} as
\begin{align*}
	\big(\hat{V}^{(t)},\hat{U}^{(t)}\big) \leftarrow \mathrm{ForceConsistency}\big(\hat{V}^{(t)},\hat{U}^{(t)}\big).
\end{align*}
In order to perform truncation, the initial estimation $\hat{U}^{(0)}$ needs to have a correct sign,
which is obtained by defining
\begin{align}
	VecSign(U) \equiv 
	\begin{cases}
	+1 &: \sum_{i=1}^{n}sign(U_{i}) \geq 0\\
	-1 &: \text{ Else  }
	\end{cases}.
\end{align}
Multiplying $\hat{U}^{(0)}$ by $VecSign(\hat{U}^{(0)})$ assures the correct sign.
%Another change to algorithm \ref{alg:NoiselessLRMC} is the removal of the normalization step as it interfere with the truncation steps.

\section{Algorithm for Noisy Data}
The improved version of Algorithm~\ref{alg:NoiselessLRMC} that includes all the changes
discussed in the previous section appears in Algorithm \ref{alg:NoisyLRMC}. It better handles noise in
the initial matrix $M$ caused by limited pairwise comparisons.
\begin{algorithm} [H]
\caption{Noisy \textsc{MC-MLE} Ranking}
\label{alg:NoisyLRMC}
\begin{algorithmic} [1]
\small
\Procedure{RankingMCMLE}{$P_{\mathcal{E}}(\mathbf{Y}),\mathcal{E},\mathbf{L},R_{Max},\Delta\omega_{Min},\delta$}
\State $\mathbf{Truncate}$: $\hat{y}_{j/i}$ based on $\mathbf{Y}_{ji}$ and Eq. \eqref{eqn:ObservationTruncation}.
\State $\mathbf{Set}$: $M_{ij}=(1/\hat{y}_{j/i}-1)$ and $M_{ii}=1$
\State $\mathbf{Set}$: $T = round\big(4 \cdot ln\big(n / (2\cdot\Delta\omega_{Min})\big) / ln(16)\big)$
\State $\mathbf{Set}$: $\hat{U}^{(0)}$ as the top left singular vector of $\frac{1}{p_{obs}}\cdot P_{\mathcal{E}}(M)$
\State $\mathbf{Clipping}$: set $\hat{U}^{(0)}_{i}$ where $|\hat{U}^{(0)}_{i}|>2 \cdot R_{Max}$ to zero
\State $\mathbf{Fix\text{ }Sign}$: $\hat{U}^{(0)} = \hat{U}^{(0)} \cdot VecSign(\hat{U}^{(0)})$
\State $\mathbf{Truncate}$: set $\hat{U}^{(0)}_{i}$ to $\frac{1}{C_{R}\cdot R_{Max}}$ if $\hat{U}^{(0)}_{i}< \frac{1}{C_{R}\cdot R_{Max}}$ \label{lst:line:Trunc1}
\State $\mathbf{Normalization}$: normalize $\hat{U}^{(0)}$ to $\hat{U}^{(0)} / \|\hat{U}^{(0)}\|_{2}$
\State $\mathbf{Set}$: $M^{(0)}=\mathbf{0}\in\mathbb{R}^{nxn}$ and $M^{(1)}= \hat{U}^{(0)}\times(1/\hat{U}^{(0)})^{\dagger}$
\State $\mathbf{Set}$: $t=1$
\While{$(t \leq T)\text{ and }(\|M^{(t)}-M^{(t-1)}\|_{F}\geq\delta)$}
	\State $\hat{V}_{q}^{(t)}=FactorMLE(M,\mathcal{E},\hat{U}^{(t-1)},\Delta\omega_{Min},\mathbf{L})$ \label{lst:line:MinV}
	\State $\hat{U}_{q}^{(t)}=FactorMLE(M,\mathcal{E},\hat{V}^{(t)},\Delta\omega_{Min},\mathbf{L})$ \label{lst:line:MinU}
	\State set $\hat{U}^{(t)}_{i}$ where $\hat{U}^{(t)}_{i}< \frac{1}{C_{R}\cdot R_{Max}}$ to $\frac{1}{C_{R}\cdot R_{Max}}$ \label{lst:line:Trunc2}
	\State set $\hat{V}^{(t)}_{i}$ where $\hat{V}^{(t)}_{i}> C_{R}\cdot R_{Max}$ to $C_{R}\cdot R_{Max}$ \label{lst:line:Trunc3}
	\State$\big(\hat{V}^{(t)},\hat{U}^{(t)}\big) \leftarrow \mathrm{ForceConsistency}\big(\hat{V}^{(t)},\hat{U}^{(t)}\big)$ \label{lst:line:ForceC}
	\State $M^{(t+1)}= \hat{U}^{(t)}\times(\hat{V}^{(t)})^{\dagger}$
	\State $t = t+1$		
	\small
\EndWhile
\State $\hat{U}^{(t-1)} = \hat{U}^{(t-1)} / \max_{i}\Big(\hat{U}^{(t-1)}_{i}\Big)$
\State \textbf{return}  $\hat{U}^{(t-1)}$
\EndProcedure
\normalsize
\end{algorithmic} 
\end{algorithm} 

The main difference between Algorithm \ref{alg:NoisyLRMC} and Algorithm \ref{alg:NoiselessLRMC} are 
lines~\ref{lst:line:MinV}\&\ref{lst:line:MinU} of Algorithm \ref{alg:NoisyLRMC}, where we replaced the Frobenius  norm minimization with the MLE (maximum likelihood estimator)
of each factor given the previously estimated factor (either $U$ or $V$) and the matrix $M$. The MLE takes into account the probabilities of all
possible values in $M$ and therefore it does not need the estimates of $M$ to be unbiased as is the case with the Frobenius norm minimization,
which minimizes the error around these entries. 

Another difference between the algorithms is the forcing of the solution in each iteration to be consistent with the known
optimal form of the solution for $U$ and $V$ (line~\ref{lst:line:ForceC} of Algorithm \ref{alg:NoisyLRMC}). This reduces the effect of errors in the values of these factors at initial iterations. Because this step may not be helpful in the case of constant $L_{ij}=L\forall (i,j)$ and $L\rightarrow\infty$ where the entries in $M$ are exact, it may happen that forcing consistency at an early
stage will slow down the convergence of the algorithm. Another difference is the truncation steps (lines~\ref{lst:line:Trunc1} , \ref{lst:line:Trunc2}\&\ref{lst:line:Trunc3} of Algorithm \ref{alg:NoisyLRMC}) added to ensure that spurious values at early iterations do not interfere with the convergence of the algorithm. Clearly, these steps are also unnecessary in the case of $L\rightarrow\infty$.

$\mathbf{Estimating\text{ }R_{Max}.}$
Under certain assumptions, if $R_{Max}$ is unknown, it can be estimated from the observation matrix $\mathbf{Y}$.
If we assume that the preference scores $\overrightarrow{\omega}$ are uniformly distributed in the range $[\omega_{Min},1]$,
then if we calculate the probability of the weakest item (corresponding to $\omega_{Min}$) to win $k$ times against a random item, we
get from the law of total probability that
\begin{align*}
	p(k|\omega_{Min},L) = \int\limits_{\omega_{Min}}^{1} \frac{1}{1-\omega_{Min}}\cdot p(k|\omega,\omega_{Min},L) \cdot d\omega.
\end{align*}
Now, we can insert the probability that the weakest item will win $k$ times against an item with a preference score $\omega$, which is simply
the binomial distribution with a probability $p=\omega_{Min}/(\omega_{Min}+\omega)$ and $L$ games
\footnotesize
\begin{align*}
	p(k|\omega_{Min},L) = \frac{\binom{L}{k}}{1-\omega_{Min}}\int\limits_{\omega_{Min}}^{1} \bigg(\frac{\omega_{Min}}{\omega_{Min}+\omega}\bigg)^{k} \bigg(\frac{\omega}{\omega_{Min}+\omega}\bigg)^{L-k} d\omega.
\end{align*}
\normalsize
Using the variable change $x = \omega / \omega_{Min}$, we get
\begin{align*}
	p(k|\omega_{Min},L) = \binom{L}{k}\cdot\frac{\omega_{Min}}{1-\omega_{Min}}\int\limits_{1}^{\omega_{Min}^{-1}} \bigg(\frac{x}{x+1}\bigg)^{L} \cdot \frac{1}{x^{k}} \cdot dx.
\end{align*}
By defining the constant
\begin{align*}
	A \equiv \frac{\omega_{Min}}{1-\omega_{Min}},
\end{align*}
and taking the expectation with respect to $k$
\begin{align}
	E[k] = A\cdot \sum_{k=0}^{L}\binom{L}{k}\int\limits_{1}^{\omega_{Min}^{-1}} \bigg(\frac{x}{x+1}\bigg)^{L} \cdot \frac{k}{x^{k}} \cdot dx.
\end{align}
Swapping the integral and sum we have
\begin{align}
	E[k] = A\cdot \int\limits_{1}^{\omega_{Min}^{-1}} \bigg(\frac{x}{x+1}\bigg)^{L} \cdot  \sum_{k=0}^{L}\binom{L}{k} \frac{k}{x^{k}} \cdot dx.
\end{align}
Solving for the sum and placing the result we have
\begin{align}
	E[k] = L\cdot A\cdot \int\limits_{1}^{\omega_{Min}^{-1}} \bigg(\frac{x}{x+1}\bigg)^{L} \bigg(\frac{x}{x+1}\bigg)^{-L} \cdot \frac{1}{1+x}  \cdot dx.
\end{align}
By calculating the integral and dividing by $L$ we get:
\begin{align}
	E\bigg[\frac{k}{L}\bigg] = \frac{\omega_{Min}}{1-\omega_{Min}} \cdot \ln\bigg(\frac{1+\omega_{Min}}{2\cdot \omega_{Min}}\bigg).
\end{align}
Notice that each entry in $\mathbf{Y}$ is a proxy of $k_{ij}/L$ where $k_{ij}$ is the number of times item $i$ won a match with
item $j$. Thus, the average value of entries in a certain row in $\mathbf{Y}$ (excluding $\mathbf{Y}_{ii}$) is simply an estimate for $E\big[\frac{k}{L}\big]$
for the $i^{th}$ item, which we will refer to as
\begin{align}
	\hat{E}_{i} = \frac{1}{|(i,j)\in\mathcal{E} \& j \neq i|} \cdot \sum_{j:(i,j)\in\mathcal{E} \& j \neq i} \mathbf{Y}_{ij}.
\end{align}
Since we want the expectation of the row corresponding to $\omega_{Min}$  (weakest item), we use the minimal value across items
\begin{align}
	\hat{E} = \min_{i}(\hat{E}_{i}).
\end{align}
We define the strictly monotonic (for $Z\in[0,1]$) function
\begin{align}
	g(Z)=\frac{Z}{1-Z} \cdot \ln\bigg(\frac{1+Z}{2\cdot Z}\bigg) - \hat{E}, \label{eqn: wMinEstimate}
\end{align}
for which the root gives us the estimation for $\omega_{Min}$ since it represent the preference score that best explains the expected number of wins for
the weakest item in the group. The root of \eqref{eqn: wMinEstimate} is easy to find up to an arbitrary required precision $\delta_{Z}$ in $O(log(1/\delta_{Z}))$ steps. After finding the root (there is only one) of the function $g(Z)$, which we denote as $Z_{0}$, the estimation for the ratio $R_{Max}$, for our selected normalization of 
$\omega_{Max}=1$, is given based on the definition in \eqref{eqn: RMaxDef}  by the ratio
\begin{align}
	\hat{R}_{Max}=\frac{1}{Z_{0}}.
\end{align}

\section{Experiments}

\subsection{Comparing the noiseless LRMC and noisy MC-MLE}

In this section we test the contributions of the modification presented in Algorithm \ref{alg:NoisyLRMC} to the straight forward adaptation 
(Algorithm \ref{alg:NoiselessLRMC}) of the {\em LRMC} algorithm presented in \cite{AltMinLRMC} in two cases. 

The first case is a scan over values of $L$ for different values of $p_{osb}$.
{\em MC-MLE} represent the final algorithm as presented in Algorithm \ref{alg:NoisyLRMC} and 
{\em LRMC} represent the noiseless algorithm as presented in Algorithm \ref{alg:NoiselessLRMC}.
A 95\% confidence bound for the rank error is calculated by fitting a generalized linear regression for binomial
distribution to the empirical rank error CDF (Monte-Carlo) calculated over many iterations.

To test the modifications, 500 iterations were used for each value of $L$ and $p_{obs}$.
The vector $\overrightarrow{\omega}$ has $N_{T}$ preference scores which always include two values $R_{max}^{-1}$ and $1$.
The rest of the $N_{T}-2$ preference scores are randomized, at each iteration, using a uniform distribution $U(0,1)$.
The uniform distribution yields a random vector $\bar{u}$ of size $N_{T}-2$, then by applying
\begin{align}
	\overrightarrow{\omega}_{i} = R_{max}^{-1} +(1-R_{max}^{-1})\cdot \bar{u}_{i}/(\max(\bar{u})-\min(\bar{u})), 
\end{align}
we get the remaining $N_{T}-2$ values of the preference scores.

\begin{figure} [H]
  \centering
  \includegraphics[width=.8\linewidth,height=5.5cm]{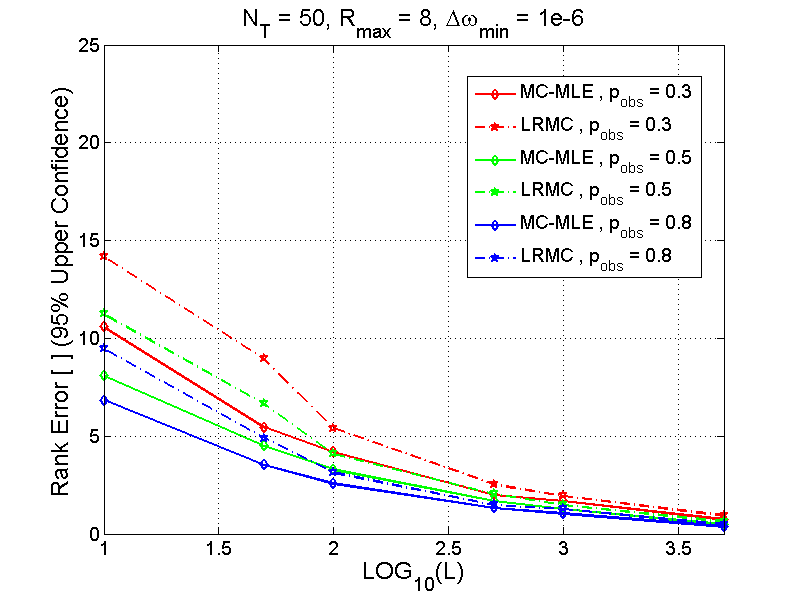}
  \caption{Scanning over different number of repeated comparisons $L$}
  \label{fig: LScan}
\end{figure}

In Fig. \ref{fig: LScan} a value of $ R_{max} = 8$  is fixed for all $L$ values and the number of preference scores ($N_{T}$) is set to 50.
The value of the constant  $C_{R}$ is chosen to be $1.4$ and we set $\delta=\Delta\omega_{min}/(20\cdot N_{T})$ to assure sufficient accuracy.

The second case scans over values of $p_{obs}$ without noise ($L \to \infty$) for different values of $R_{max}=\omega_{max}/\omega_{min}$.
Because the 95\% confidence bound on the rank error appears to be very small for this scenario, the probability of a ranking error
is presented instead. This case is tested for several values of $N_{T}$ (the number of items). In this case, there is no need for
truncation so $C_{R}$ is chosen to be $1000$ such that the truncation has no effect. Another change is that the forcing of the factors to the form
of the optimal solution at the initial iterations causes the algorithm to converge to a small error larger than zero for all $p_{obs}$ values.
Therefore, when $L\rightarrow\infty$ we remove line \ref{lst:line:ForceC} from Algorithm \ref{alg:NoisyLRMC}.
\begin{figure} [H]
  \centering
  \includegraphics[width=.95\linewidth,height=7.0cm]{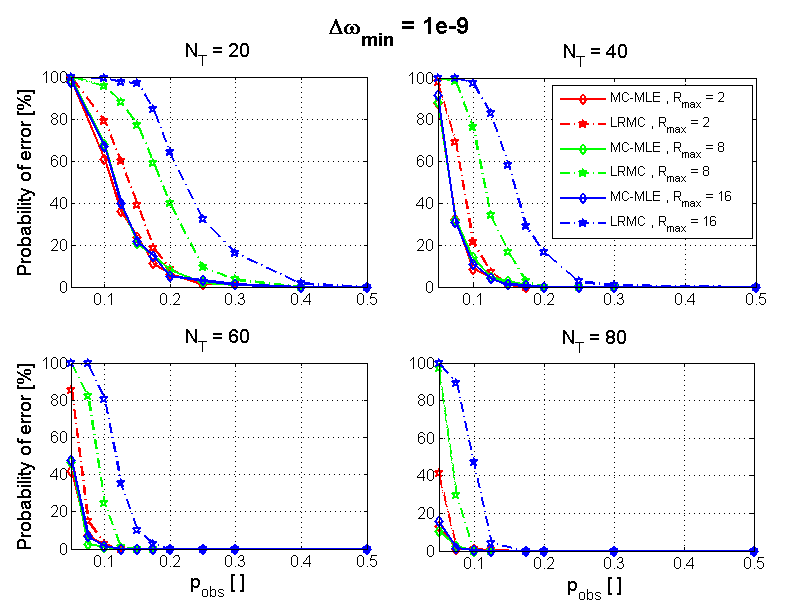}
  \caption{Scanning over different observation probabilities $p_{obs}$}
  \label{fig: pScan}
\end{figure} \noindent
As can be seen in Fig. \ref{fig: pScan} the {\em MC-MLE} ranking algorithm shows significant improvement (over the simple adaptation of {\em LRMC}) for low values of $p_{obs}$
even in the simple completion problem without noise. The improvement becomes more significant as higher values of $R_{max}$ are used. In Fig. \ref{fig: LScan} we can
see that the modified algorithm {\em MC-MLE} performs better than the simple {\em LRMC} on all value of $L$ and for all values of $p_{obs}$ tested. The difference between 
the algorithms becomes smaller as the value of $L$ increases. However, as can be seen in Fig. \ref{fig: pScan} this is only true for a high enough value of $p_{obs}$. 

For both scenarios the algorithm received the value of $R_{max}$ as an input. However as we tested in the following simulations the value of $R_{max}$ can be estimated
from the data matrix $\mathbf{Y}$ instead of being inserted as input to the algorithm.

\subsection{Comparison to current state-of-the-art}
A comparison to other ranking methods is presented here. We compare {\em Rank Centrality} ({\em RC}) \cite{RC}
and {\em Spectral MLE} ({\em SMLE}) \cite{SMLE} algorithms to our suggested {\em MC-MLE} algorithm.
 
For {\em SMLE} we used the constant $c3=0.1$ instead of $c3=1$ as used in \cite{SMLE}
in order to get better results for {\em SMLE}. In this simulation 200 {\em Monte-Carlo} trials were used with $ R_{max} = 2$.
The reported results are obtained by averaging over all the {\em Monte-Carlo} trials. The preference scores ($\overrightarrow{\omega}$) are randomized uniformly
as previously described in the testing of the algorithm modifications. For all the following results the value of $R_{max}$ is estimated from the data
matrix $\mathbf{Y}$ and inserted to the {\em MC-MLE} algorithm, so it is not an input of the overall algorithm. The value of $\Delta\omega_{min}$ is $1e-6$ for
all the following simulations, and as before $\delta=\Delta\omega_{min}/(20\cdot N_{T})$. 

\begin{figure} [H]
  \centering
  \includegraphics[width=.8\linewidth,height=5.5cm]{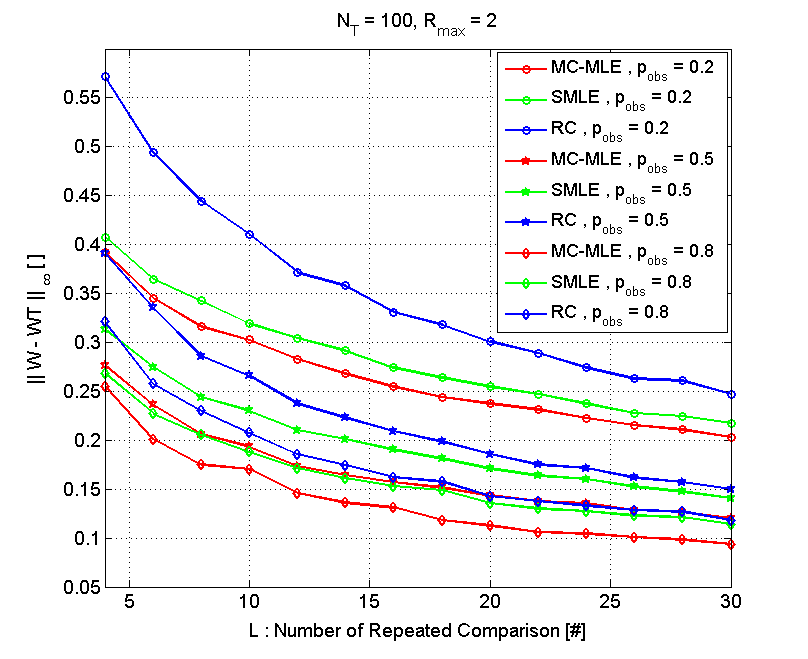}
  \caption{Scanning over different number of repeated comparisons $L$}
  \label{fig: LScanRC}
\end{figure}

\begin{figure} [H]
  \centering
  \includegraphics[width=.8\linewidth,height=5.5cm]{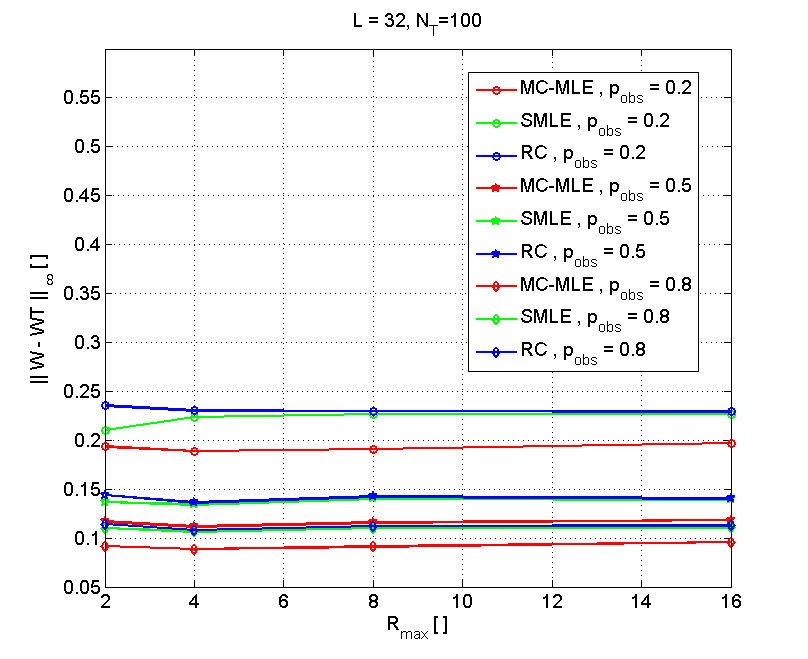}
  \caption{Scanning over different weights ratio ($R_{max}$) values}
  \label{fig: RScanRC}
\end{figure}

We observed that at low enough values of $L$, choosing a small value for $C_{R}$ may result in a too strict truncation at initialization and so a 
homogeneous vector of equal scores. Thus, the value of $C_{R}$ needs to slightly increase for larger $p_{obs}$ and $L$, but also for very low values of $L$. 
For simulation we used:
\begin{align}
	C_{R} = 
	\begin{cases}
	1.2 &: p_{obs} \leq 0.2 \\
	1.4 &: p_{obs} > 0.2 \text{ , } L \geq 10 \\
	1.8 &: p_{obs} > 0.2 \text{ , } L < 10. \\
	\end{cases}
\end{align}
As can be seen in Fig.~\ref{fig: LScanRC}, {\em MC-MLE} achieves better performance than both {\em RC} and {\em SMLE}
for all values of $L$ in this simulation. The performance difference is slightly reduced for the smaller observation probability, and perhaps
an even better choice of $C_{R}$ can improve this result as we did not optimize over this value. This improvement is consistent
over different values of $R_{max}$ as can be demonstrated in Fig. \ref{fig: RScanRC}. In this simulation a value of 100 is used
for the number of items $ N_{T}$.

Further decreasing the value of $\Delta\omega_{min}$ was tested for improved results and displayed no significant benefit so we
fixed the value of $\Delta\omega_{min}$ throughout all the following simulations.

\begin{figure} [H]
  \centering
  \includegraphics[width=.8\linewidth,height=5.5cm]{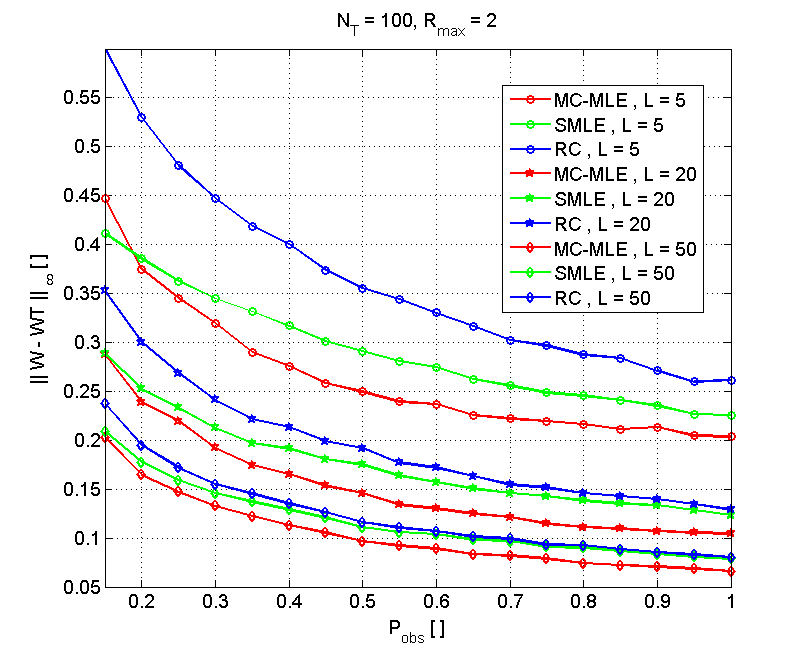}
  \caption{Scanning over different observation probabilities $p_{obs}$}
  \label{fig: pScanRC}
\end{figure}

As can be seen in Fig. \ref{fig: pScanRC} {\em MC-MLE} achieves better performance than both {\em RC} and {\em SMLE}
for all values of $p_{obs}$ in this simulation (apart from one scenario where $L=5$ and $p_{obs}=0.2$). 
Note that the performance difference increases as the number of comparisons $L$ is reduced.

In Fig. \ref{fig: RScanRC} we can observe that the performance improvement of {\em MC-MLE} compared with {\em RC} and {\em SMLE}
is independent of the choice of $R_{max}$ even though {\em SMLE} takes as input the value of $R_{max}$ and in the case of {\em MC-MLE}
it is estimated from the observation matrix $\mathbf{Y}$.

\begin{figure} [H]
  \centering
  \includegraphics[width=.8\linewidth,height=5.5cm]{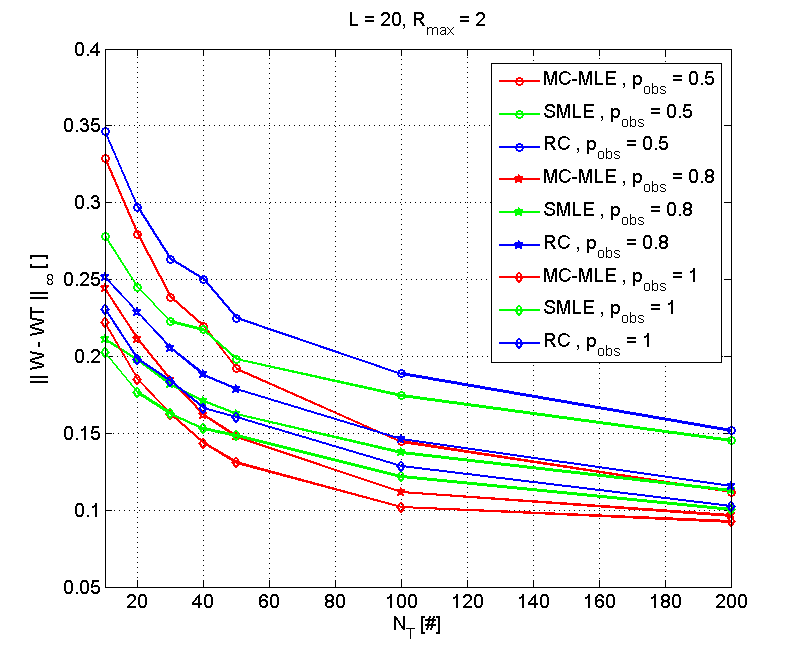}
  \caption{Scanning over different number of items $N_{T}$}
  \label{fig: NScanRC}
\end{figure}

In Fig. \ref{fig: NScanRC} we can see that the performance of {\em MC-MLE} depends strongly on $N_{T}$ and that
as the value of $p_{obs}$ decreases a larger value of $N_{T}$ is required to achieve better performance than {\em SMLE}
for a constant value of $L=20$ and $R_{max}=2$. Perhaps an optimization of $C_{R}$ could also help in this case. 

\subsection{Evaluation on weather data-base}
The LRMC ranking algorithm was tested on a weather data-base of monthly measurements from 45 states over several years
starting from July 2001. For each pair of states in each month a match was performed by adding a score of 1 point for the state
with the higher temperature and a score of 0.5 to both in cases of ties. The probability estimator $y_{i/j}$ was calculated by normalizing each
total score by the total amount of points given.

\begin{figure} [H]
  \centering
  \includegraphics[width=.8\linewidth,height=5.2cm]{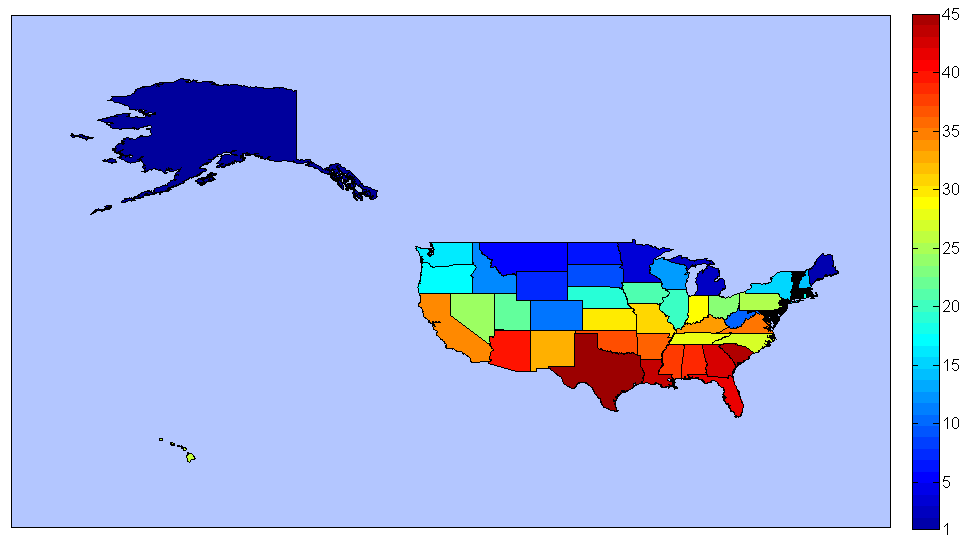}
  \caption{State ranked from coldest (Blue) to warmest (Red) with $p_{obs} = 0.7$}
  \label{fig: StateRank}
\end{figure} 

The states Connecticut, Delaware, Maryland, Massachusetts, New Jersey, Vermont (and Washington DC) were not ranked, as
they did not appear in the data base and had no data. Therefore, they appear in black on the map in Fig.~\ref{fig: StateRank}.

\begin{figure} [H]
  \centering
  \includegraphics[width=.8\linewidth,height=4.5cm]{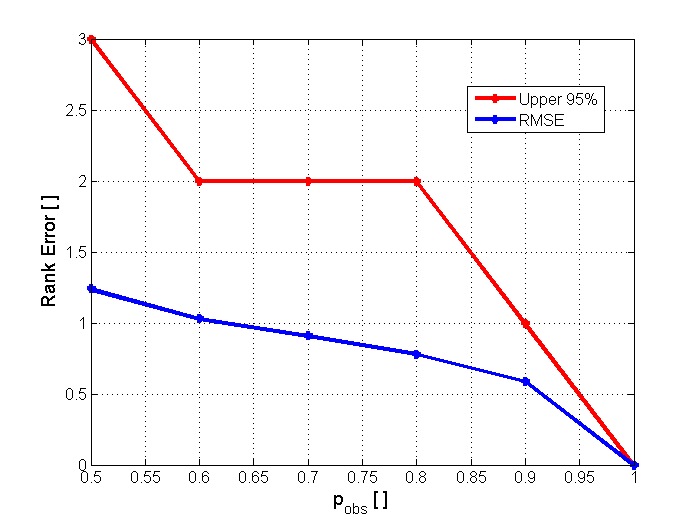}
  \caption{Weather ranking error}
  \label{fig: StateRankError}
\end{figure}

In Fig. \ref{fig: StateRankError} the ranking accuracy is presented for different observation probabilities $p_{obs}$. For this simulation, in each iteration, the data matrix containing the ratio estimations $M$ was randomly obscured using the operator $P_{\mathcal{E}}(X)$ with a $p_{obs}$ probability of {i,j} belonging to $\mathcal{E}$.
This meant discarding a subset of comparison data between certain states. In each iteration the overall ranking between all the states was then calculated using {\em MC-MLE}.
After 30 iteration the stability of the ranking was measured as the root mean square error (RMSE) of the rank of all states compared with the rank based on $p_{obs} = 1$.
In addition, the 95\% confidence interval on the rank error (compared with $p_{obs} = 1$) was calculated and the upper limit is displayed in Fig. \ref{fig: StateRankError}.

From this figure it can be concluded that a stable ranking is achieved, since the RMSE is relatively small (less than 2) for observation probabilities as low as $0.5$.
In addition, the ranking appears to be accurate since the 95\% confidence bound on the ranking error (compared with  $p_{obs} = 1$) is equal to 3.0 (which is relatively small) for observation probabilities as low as $0.5$. 
We observed that a state that had a larger deviation (compared to the other states) for a certain observation maintained a larger deviation for other observation probabilities. 
This is due to the fact that states that are similar (in terms of their weather) are harder to rank for all observation probabilities.

\subsection{Analysis of the football (soccer) data}
\label{sec:football}

The {\em MC-MLE} ranking algorithm was also applied to football matches data using the Olympic, European
championship (UEFA) and FIFA world cup games. 
The probability estimator for $i$ over $j$ was calculated by giving 3 points for each win of $i$ over $j$ 
regardless of the match score and one point to both $i$ and $j$  in cases of a tie score. The sum of the points for both $i$ and $j$ are then normalized by 
the total amount of points distributed to both. In order to compare with the FIFA ranking algorithm, the FIFA ranking was taken for the top 50 teams of each year from 2008 to 2016.
The {\em MC-MLE} ranking was constructed for each year on the top 50 teams from the FIFA ranking of December of the same year using the data of the previous years. The number 
of previous years taken into the estimation was tested on another time window (years 1999-2008) to find an optimal window size that includes enough data to properly compare all teams but at the same time does not take into account too old matches that may be irrelevant to the teams' current status. 

To compare the algorithms, for each year, a score was given to each algorithm by looking at all the games of that year
(involving the FIFA top 50 teams). For every game that ended with a team winning, each algorithm that ranked the winning team higher than the losing team
had received a point. For each tie, the algorithm that ranked the teams closer (than the other algorithm) had received half a point. In tie cases where the distance in ranking is identical
for both algorithms, each received half a point. For each year the {\em MC-MLE} algorithm ranked the top 50 teams from FIFA's ranking of December of the previous year using data from previous years. 

\begin{table*}
\centering
\caption{Scores for FIFA and LRMC ranking comparison. }
\begin{tabular}{l | rrrrrrrrrr}
Method & 2017 & 2016 & 2015 & 2014 & 2013 & 2012 & 2011 & 2010 & 2009 & 2008\\
MC-MLE 	& \textbf{22.5}	& \textbf{67.5}	& \textbf{24.5}	& 43.5                	& \textbf{45.0}	& \textbf{45.5}	& \textbf{39.0}	& \textbf{44.0}	& \textbf{39.5} & \textbf{38.5} \\
FIFA		& 19.5               	& 67.0               	& 22.0		& \textbf{50.0}	& 42.5                	& 44.0               	& 36.0               	& 39.0                	& 38.0               & 37.5                 \\
\end{tabular} 
\end{table*} 

Though in some years the results are close, {\em MC-MLE} scored better in 9 out of the 10 years tested,
tied in one out of 10 years and scored lower in one out of 10 years. The optimal window size chosen is 8 years of backward data.
If we use a larger window of 9 years {\em MC-MLE} has a higher score on 6 years and two tie scores out of 10 years,
and if we use a smaller window of 7 years {\em MC-MLE} has a higher score on 7 years and two tied scores out of 10 years.
The choice of the window is important. However, the {\em MC-MLE} algorithm is still better on a smaller and larger choice of window.
On the validation data (years 1999-2008), the {\em MC-MLE} performed better in the prediction of the results of 7 years and tied on one year based on the 8 years window size.
The {\em MC-MLE} algorithm also performed better (on the validation set) when a window size of 7 or 9 years is applied. The optimal window
on the validation is 9 years of backward look having better results than FIFA's method of ranking on 8 of the 10 years tested.

\section{Conclusions}
In this paper, we proposed a new method for rank recovery based on a matrix completion approach. We presented
a strategy for rank recovery from partial observations that supports limited comparisons,
which introduces noise to the matrix completion model. The proposed approach was tested in both a limited comparisons scenario and
a noiseless scenario and had shown improvement over the state-of-the-art. We tested the stability of our proposed method under different observation
probabilities on a weather data-set assembled from limited comparisons. Finally, we compared  FIFA's ranking to our ranking and evaluated them by testing
their performance in predicting match results on successive years (for ten years). Our solution achieved better performance than FIFA's method for team ranking.

Our matrix completion approach can also be adapted to other models. For instance to the model proposed in \cite{GeneralizedBTL}:
\begin{gather*}
   	P(\text{i beats j})=\omega_i/(\omega_i+\theta\cdot\omega_j),\\
	P(\text{j beats i})=\omega_j/(\theta\cdot\omega_i+\omega_j),\\
	P(\text{i ties j})=(\theta^{2}-1)\cdot\omega_i\omega_j/[(\omega_i+\theta\cdot\omega_j)(\theta\cdot\omega_i+\omega_j)],
\end{gather*}
which includes tie results. for this model we construct the incomplete low-rank matrix as:
\begin{gather*}
   	M_{ij}=1/P(\text{j beats i})-1=\theta\cdot\frac{\omega_{i}}{\omega_{j}}\\
	M_{ji}=1/P(\text{i beats j})-1=\theta\cdot\frac{\omega_{j}}{\omega_{i}}
\end{gather*}
This is equivalent to the original problem we defined in this paper up to a scale (which does not effect our solution).
After recovering the preference scores vector $\vec{\omega}$, we may use all the tie probabilities estimators and
the preference scores estimation to recover the factor $\theta$. We leave further analysis of this model to a future work.

\section*{Acknowledgment}
This work is partially supported by ERC-StG SPADE PI Giryes.

%\pnasbreak 
% If you see unexpected formatting errors, try commenting out this line
% as it can run into problems with floats and footnotes on the final page.
%\pnasbreak

% Bibliography

%\bibliography{pnas-bib}

\end{document}